\documentclass[conference]{IEEEtran}

\usepackage{svg}
\usepackage{array}
\usepackage{xcolor}
\usepackage{amsmath}
\usepackage{comment}
\usepackage{siunitx}
\usepackage{listings}
\usepackage{amsfonts}
\usepackage{textcomp}
\usepackage{booktabs}
\usepackage{tabularx}
\usepackage{graphicx}
\usepackage{subfiles}
\usepackage{booktabs}
\usepackage{setspace}
\usepackage{cleveref}
\usepackage{pdfpages}
\usepackage{subcaption}
\usepackage{algpseudocode}
\usepackage{float}
\usepackage[normalem]{ulem}
\usepackage[scaled]{beramono}
\usepackage[linesnumbered,ruled,vlined]{algorithm2e}

\def\BibTeX{{\rm B\kern-.05em{\sc i\kern-.025em b}\kern-.08emT\kern-.1667em\lower.7ex\hbox{E}\kern-.125emX}}

\newcolumntype{Y}{>{\centering\arraybackslash}X}
\newcolumntype{Z}{>{\centering\arraybackslash\hsize=.4\hsize}X}
\newcolumntype{A}{>{\centering\arraybackslash\hsize=.65\hsize}X}
\newcolumntype{B}{>{\centering\arraybackslash\hsize=1.15\hsize}X}

\makeatletter
\newcommand{\linebreakand}
{
  \end{@IEEEauthorhalign}
  \hfill\mbox{}\par
  \mbox{}\hfill\begin{@IEEEauthorhalign}
}
\def\ALG@special@indent{%
    \ifdim\ALG@thistlm=0pt\relax
        \hskip-\leftmargin
    \else
        \hskip\ALG@thistlm
    \fi
}

\begin{document}

\title{Machine Learning-Based Estimation Of Wave Direction For Unmanned Surface Vehicles}

\author
{
\IEEEauthorblockN{Manele Ait Habouche}
\IEEEauthorblockA{
\textit{Université de Bretagne Occidentale} \\
\textit{Manele.AitHabouche@univ-brest.fr}}
\and
\IEEEauthorblockN{Mickaël Kerboeuf}
\IEEEauthorblockA{
\textit{Université de Bretagne Occidentale} \\
\textit{Mickael.Kerboeuf@univ-brest.fr}}
\and
\IEEEauthorblockN{Goulven Guillou}
\IEEEauthorblockA{
\textit{Université de Bretagne Occidentale} \\
\textit{Goulven.Guillou@univ-brest.fr}}
\and
\linebreakand
\IEEEauthorblockN{Jean-Philippe Babau}
\IEEEauthorblockA{
\textit{Université de Bretagne Occidentale} \\
\textit{Jean-Philippe.Babau@univ-brest.fr}}
}

\maketitle

\begin{abstract}
Unmanned Surface Vehicles (USVs) have become critical tools for marine exploration, environmental monitoring, and autonomous navigation. Accurate estimation of wave direction is essential for improving USV navigation and ensuring operational safety, but traditional methods often suffer from high costs and limited spatial resolution. This paper proposes a machine learning-based approach leveraging LSTM (Long Short-Term Memory) networks to predict wave direction using sensor data collected from USVs. Experimental results show the capability of the LSTM model to learn temporal dependencies and provide accurate predictions, outperforming simpler baselines.
\end{abstract}

\section{Introduction}
\label{introduction}
Unmanned Surface Vehicles (USVs) are gaining prominence across various applications such as marine exploration, environmental monitoring, and autonomous navigation. A significant challenge for USVs is accurately estimating wave characteristics, particularly wave direction.

The estimation of wave direction is a complex problem due to the dynamic and unpredictable nature of ocean waves. Traditional methods, such as wave buoys, satellites, and wave radars, are often constrained by high costs and limited spatial resolution \cite{cheng_noveldenselyconnected_2020}. Machine learning offers a promising alternative, as it does not depend on pre-defined mathematical models, which may not perform well under varying wave conditions \cite{cheng_noveldenselyconnected_2020}. This approach enables the prediction of wave characteristics directly from USV sensor data. Such capability is crucial for improving the safety and operational efficiency of USVs and minimizing risks during critical wave encounters \cite{mittendorf_seastateidentification_2022}.

In this context, this paper introduces a machine learning model that leverages LSTM (Long Short-Term Memory) networks to estimate wave direction from USV sensor data. This task is treated as a regression problem. The model's performance is assessed using real-world datasets, and the results indicate that our model outperforms simpler baselines.

The remainder of this paper is organized as follows. Section \ref{model_overview} provides an overview of our approach, while Section \ref{evaluation} evaluates the model's performance. Section \ref{related_work} presents related work, and Section \ref{conclusion} concludes the paper and opens some perspectives.

\section{Proposed model}
\label{model_overview}
\begin{figure*}
    \centering
    \includegraphics[width=0.75\linewidth]{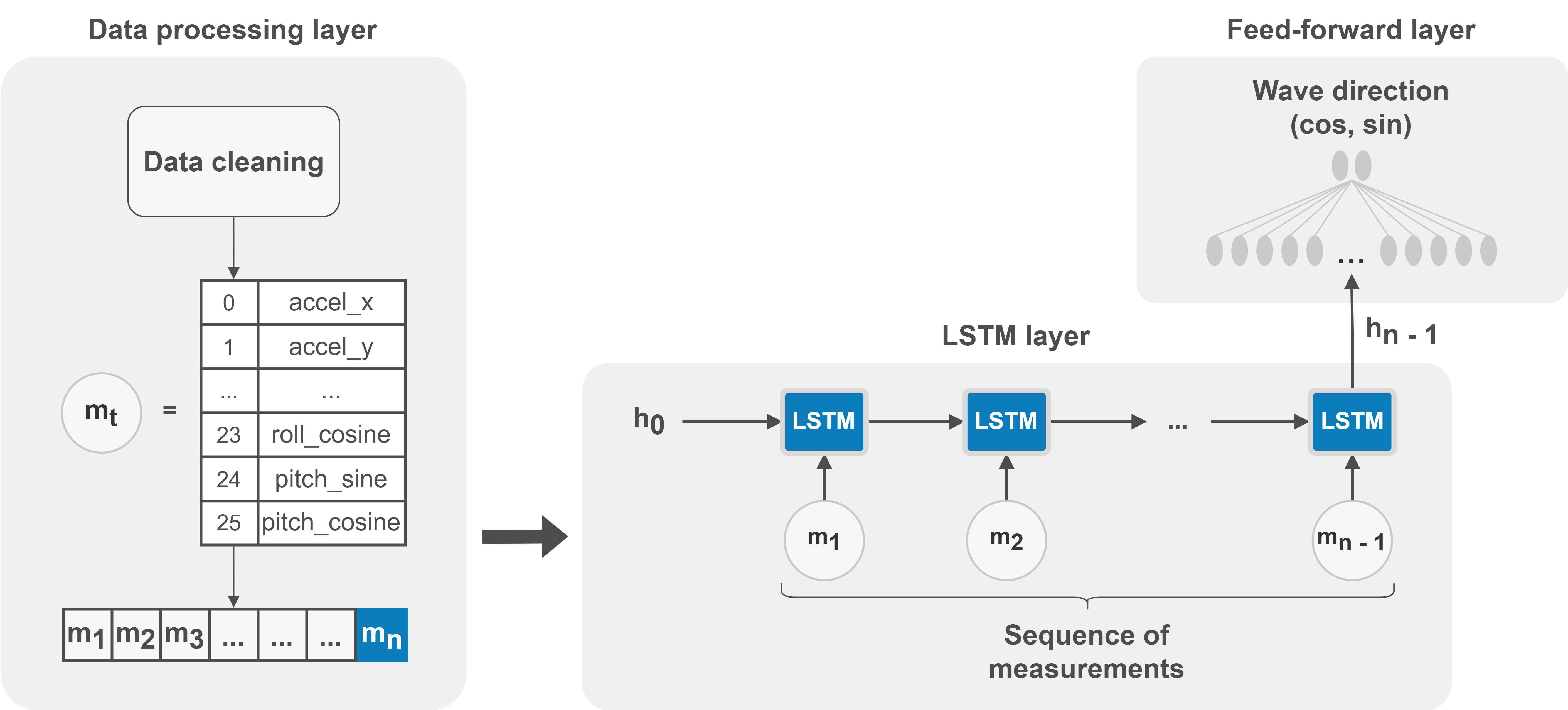}
    \caption{Model overview - A temporal view}
    \label{fig:model}
\end{figure*}

Our model leverages an LSTM (Long Short-Term Memory) \cite{hochreiter_longshorttermmemory_1997} network to predict wave direction based on the drone's navigation and orientation data. RNNs (Recurrent Neural Network) are neural network architectures specifically designed to process data sequences of variable lengths. LSTMs are a type of RNNs, capable of learning long-term dependencies. This is achieved through gating mechanisms that regulate the flow of information, enabling the network to retain or discard particular sequential information.

\subsection{Data processing}
\label{model_overview_data_processing}

The data collected from the drone's IMU (Inertial Measurement Unit) undergoes the following pre-processing steps.

\subsubsection{Data cleaning and segmentation into transects}
Noisy and irrelevant sensor readings, as well as data from periods when the drone is stationary or otherwise not relevant, are filtered out. This ensures that only meaningful data are retained. The cleaned data are then segmented into transects based on the drone's trajectory. This process is semi-automatic.
Additionally, angular measurements such as roll, pitch, and yaw can exhibit discontinuities due to their cyclical nature (e.g., transitioning from \(\pi\) to \(- \pi\)). To address these issues, these measurements are transformed into their sine and cosine components. The resulting features are detailed in Table \ref{tab:features}.

\subsubsection{Sequence generation}
To effectively capture the temporal dependencies in the drone's movement, which is significantly influenced by wave dynamics, the time series data of each feature are segmented into fixed-length sequences or windows. Each sequence includes \(n\) consecutive time steps.

\subsubsection{Standardization and train/test split}
Given that the IMU measures different physical quantities with varying ranges and units, data need to be scaled to remove any potential biases. The standard score of a sample \(s\) for each feature \(i\) is computed as follows:

\[s_{i}' = \frac{s_{i} - \mu_{i}}{\sigma_{i}}\]

Here, \(\mu_{i}\) represents the mean and \(\sigma_{i}\) the standard deviation of the training samples for feature \(i\). Following standardization, the data are divided into training and testing datasets. It is important to note that standardization is performed separately on the training and testing datasets to prevent data leakage, which could otherwise allow information from the testing dataset to influence the training process.

\begin{table}[ht]
\centering
\setlength{\tabcolsep}{1pt}
\begin{tabularx}{\linewidth}{|Z|A|}
\hline
\textbf{Category} & \textbf{Features} \\
\hline
\textbf{Linear acceleration} & accel\_x, accel\_y, accel\_z \\
\hline
\textbf{Angular velocity} & gyro\_x, gyro\_y, gyro\_z \\
\hline
\textbf{Magnetic field} & mag\_x, mag\_y, mag\_z \\
\hline
\textbf{Velocity} & north\_vel, east\_vel, down\_vel \\
\hline
\textbf{Altitude} & alt \\
\hline
\textbf{Euler angles} & yaw\_sine, yaw\_cosine, roll\_sine, roll\_cosine, pitch\_sine, pitch\_cosine \\
\hline
\textbf{Quaternion orientation} & w\_quat, x\_quat, y\_quat, z\_quat \\
\hline
\textbf{Heave} & heave\_period, heave\_motion, heave\_accel \\
\hline
\end{tabularx}
\caption{Extracted sensor features for predicting wave direction}
\label{tab:features}
\end{table}

\subsection{Model architecture}
\label{model_overview_architecture}

Figure \ref{fig:model} depicts the overall architecture of our model, which consists of three layers. The first layer, the data processing layer, prepares the raw navigation and orientation data into a format suitable for the next layers. The processed data are then passed to the LSTM layer, where they are sequentially processed through multiple LSTM cells. Finally, the output from the LSTM layer is fed into the feed-forward layer, which computes the wave direction as a vector of cosine and sine values. 

The LSTM model is designed for a regression task and aims to predict the wave direction relative to the drone's movement. The data processing layer generates sequences of measurements over \(n\) time steps. However, only the first \(n - 1\) measurements are fed into the LSTM network. This enables the model to predict the wave direction for the final time step based on the preceding observations.

As illustrated in Figure \ref{fig:model}, the sequence of measurements \((m_1, m_2, \dotsc, m_{n - 1})\) is processed sequentially through a series of LSTM cells. At each time step \(t\), an LSTM cell takes the current measurement \(m_t\) and the hidden state from the previous time step \(h_{t - 1}\) as inputs. The final hidden state \(h_{n - 1}\), which encapsulates the learned information from the sequence, is then passed to a feed-forward layer. This layer maps the output of the LSTM model to the desired prediction space, namely the sine and cosine values of the wave direction.

\section{Evaluation}
\label{evaluation}
\subsection{Experiments}
\label{evaluation_experiments}

Experiments were carried out using data from a USV (Unmanned Surface Vehicle) equipped with various sensors, including an Ellipse INS (Inertial Navigation System) for orientation and navigation and a GNSS (Global Navigation Satellite System) module with RTK (Real-Time Kinematic) positioning for location tracking.

The experiments were conducted in two distinct settings to validate the model's effectiveness across both controlled and real-world conditions. 

\begin{figure*}[ht]
\centering
\begin{subfigure}{0.45\linewidth}
    \centering
    \includegraphics[width=\linewidth]{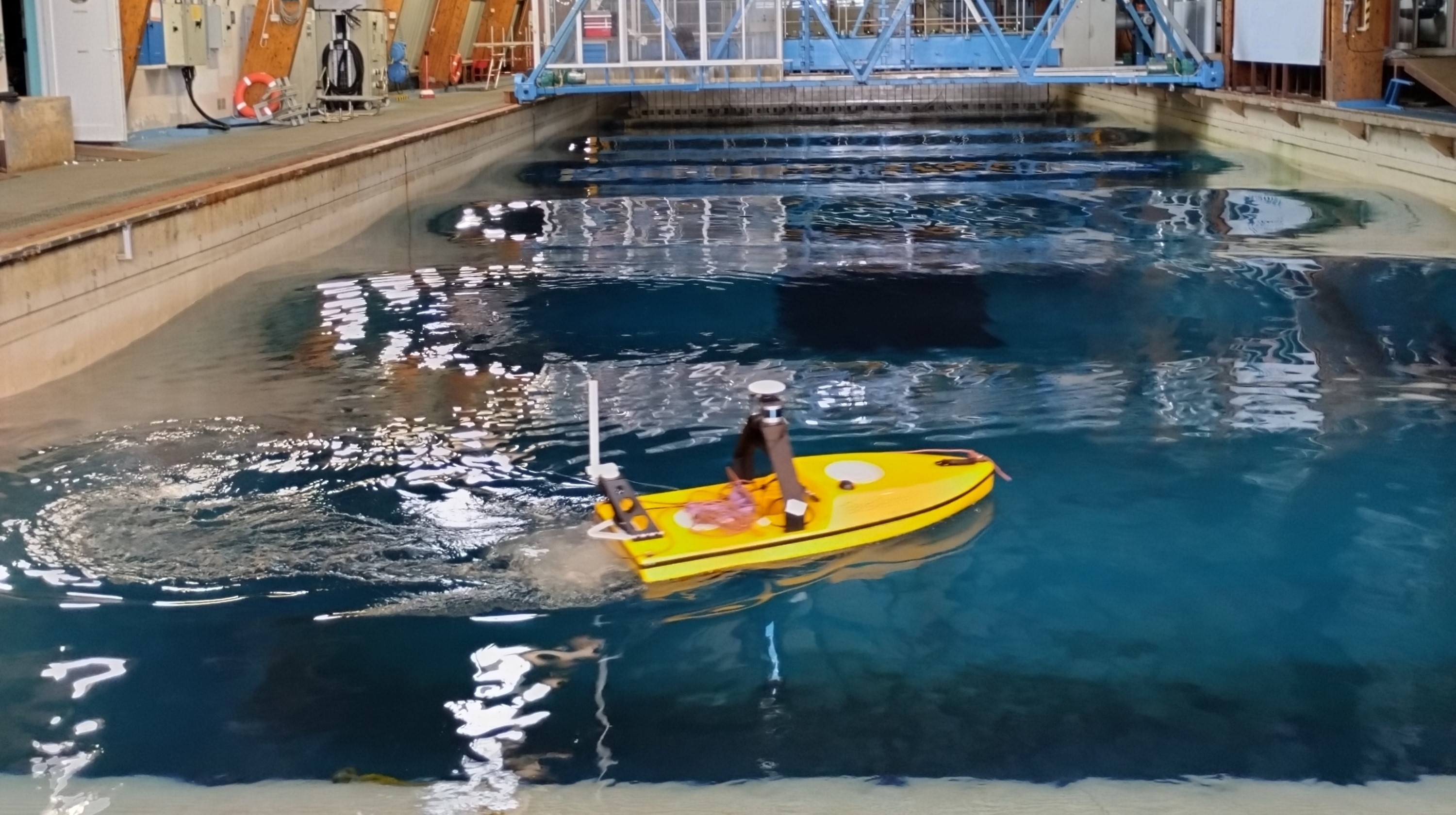}
    \caption{Experimental pool setting}
    \label{fig:pool}
\end{subfigure}
\hfill
\begin{subfigure}{0.45\linewidth}
    \centering
    \includegraphics[width=\linewidth]{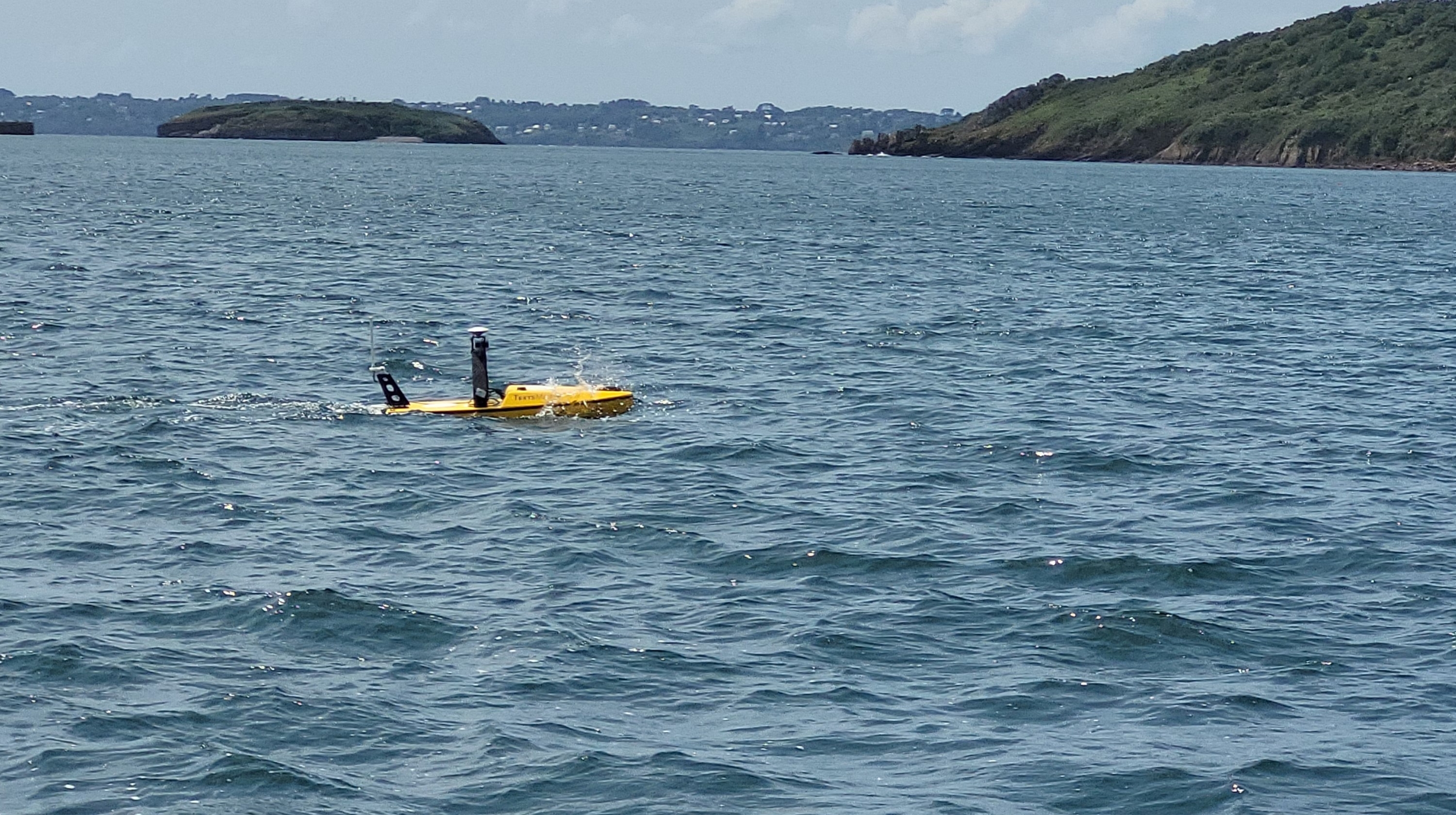}
    \caption{Open sea setting}
    \label{fig:sea}
\end{subfigure}
\caption{Experimental settings for data collection}
\label{fig:experiments}
\end{figure*}

\begin{figure}
    \centering
    \includegraphics[width=0.5\linewidth]{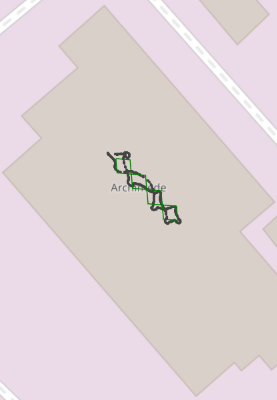}
    \caption{Drone's trajectory in experimental pool}
    \label{fig:pool_trajectory}
\end{figure}

The first setting was an experimental pool, depicted in Figure \ref{fig:pool}, that features a wave generator capable of generating controlled wave conditions with specific heights and periods. In our experiments, waves with heights ranging from 10 to 20 centimeters and periods of 2 seconds and 2.5 seconds were generated. Ten different trajectories were tested. Figure \ref{fig:pool_trajectory} illustrates an example of these trajectories.

\begin{figure}
    \centering
    \includegraphics[width=0.65\linewidth]{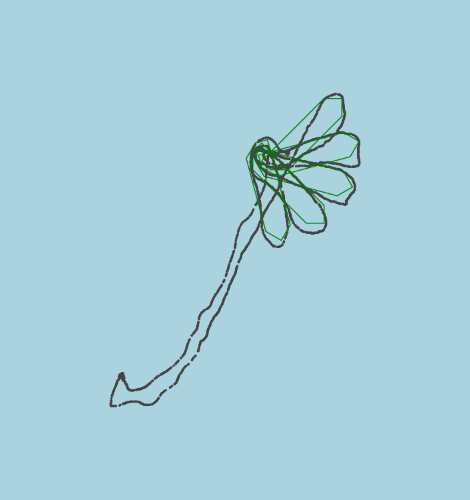}
    \caption{Drone's trajectory in open sea}
    \label{fig:sea_trajectory}
\end{figure}

The second set of experiments took place in the open sea, as depicted in Figure \ref{fig:sea}. Six trajectories following a flower pattern were tested to evaluate the drone's response from multiple orientations relative to the wave direction. An example of this trajectory is shown in Figure \ref{fig:sea_trajectory}.

In both settings, various PID values for the drone's steering rate and speed/throttle were tested to assess the model’s performance under different operational conditions.


\subsection{Determining ground truth}
\label{evaluation_ground_truth}

As detailed in Section \ref{model_overview_architecture}, the target variables are the sine and cosine components of the drone's movement relative to the dominant wave direction, which is computed by subtracting the wave direction from the drone's yaw. 

In the controlled environment of the experimental pool, the wave direction is determined by the pool's heading, since waves consistently originate from a fixed direction. 

For the open sea experiments, the wave direction at the testing site is determined using traditional maritime estimation techniques (use of fixed landmarks).

\subsection{Training and test data}
\label{evaluation_training_test}

The data from the experimental pool is used to train our model. A total of 43,319 raw data points are collected across the 10 trajectories, with each trajectory lasting approximately 2 minutes. 

Following the pre-processing steps described in Section \ref{model_overview_data_processing}, the dataset is reduced to 32,557 data points. This dataset is then split into training and testing datasets, with 80\% of the data used for training and the remaining 20\% for testing.

\subsection{Evaluated metrics}
\label{evaluation_metrics}

Two metrics are used to evaluate the model's performance.

\subsubsection{MAPE (Mean Absolute Percentage Error)}
This metric \cite{chen_assessingforecastaccuracymeasures_2004} expresses the prediction error as a percentage. Its symmetric nature is particularly appropriate in our context, where both overestimations and underestimations need to be treated with equal importance to avoid directional bias. The formula for MAPE is: 
\[
\mathit{MAPE} = \frac{1}{p} \times \sum_{i=1}^{p} \frac{\left| {\mathit{predicted}_i} - \mathit{{actual}_i} \right|}{\mathit{{actual}_i}}
\]

\subsubsection{Angular score}
To evaluate the accuracy of wave direction predictions, we define an angular score similar to RMSE (Root Mean Squared Error) but adapted for angular data, which are cyclic by nature. The predicted and actual values are first converted from their sine and cosine components into angles using the \(atan2\) function. The angular differences are then computed and normalized within a \([- \pi, \ \pi]\) range. Finally, the angular score is computed as the RMSE of these normalized differences as follows:
\[
\mathit{Angular \ score} = \sqrt{\frac{1}{p} \times \sum_{i=1}^{p} \mathit{normalized\_diff_{i}}^2}
\]
 
\subsection{Hyperparameters tuning}
\label{evaluation_tuning}

Tuning hyperparameters is a critical aspect of training recurrent neural networks, as highlighted in \cite{macneil_finetuningstabilityrecurrent_2011}, to ensure training convergence and effective generalization of models when applied to new data. However, this tuning phase is computationally expensive, limiting the number of configurations that can be practically tested.

Based on preliminary experiments, the number of training epochs is set at 50, the MSE (Mean Squared Error) is selected as the loss function, and the ADAM optimizer \cite{kingma_adammethodstochastic_2017}, widely used in deep learning, is employed as the optimization algorithm.

In this section, the effect of four hyperparameters on model performance is discussed. Table \ref{tab:tuning} summarizes the results.

\begin{itemize}
\item Sequence size. It represents the length of the input sequences fed into the LSTM network.
\item Hidden size. It represents the dimension of the hidden state in each LSTM layer.
\item Stacked LSTMs. This term describes the architecture of the LSTM network where multiple LSTM layers are stacked on top of each other.
\item Learning rate. It determines the magnitude of updates made to the model weights in response to the estimated error each time the model weights are updated.
\end{itemize}

\subsubsection{Sequence size}
\begin{figure*}[ht]
    \centering
    \begin{subfigure}[b]{0.45\textwidth}
        \centering
        \includegraphics[width=\textwidth]{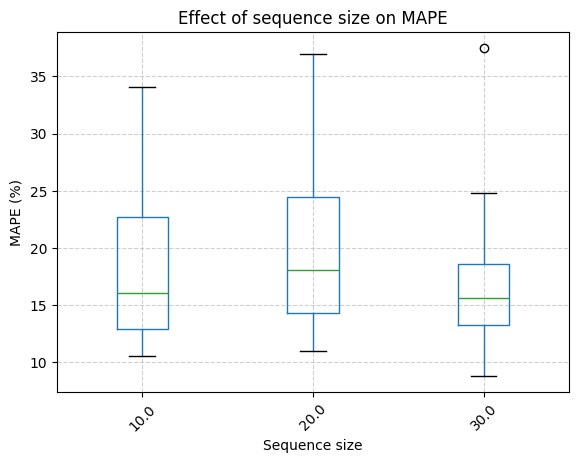}
        \caption{Sequence size effect on MAPE}
        \label{fig:sequence_size_effect_mape}
    \end{subfigure}
    \hfill
    \begin{subfigure}[b]{0.45\textwidth}
        \centering
        \includegraphics[width=\textwidth]{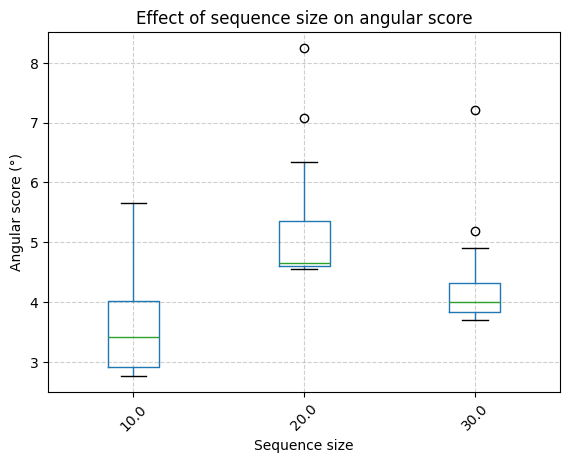}
        \caption{Sequence size effect on angular score}
        \label{fig:sequence_size_effect_angular_score}
    \end{subfigure}
    \caption{Sequence size effect on model performance}
    \label{fig:sequence_size_effect}
\end{figure*}

As illustrated in Figures \ref{fig:sequence_size_effect_mape} and \ref{fig:sequence_size_effect_angular_score}, the relationship between sequence size and model performance is not linear and appears to depend on other factors such as the number of hidden units and stacked LSTMs. While sequence lengths of 10 and 30 generally outperform a sequence length of 20, the optimal size depends on the configuration. For instance, the best angular score (2.77°) is achieved with a sequence length of 10. However, the lowest MAPE (8.83\%) is obtained with a sequence length of 30.

\subsubsection{Hidden state dimension}
Increasing the number of hidden units improves performance up to a certain threshold. Moving from 10 to 20 hidden units often reduces the MAPE and improves the angular score. However, further increasing the number to 100 does not consistently yield better results. More specifically, when the hidden state dimension is set to 100, improvements are generally observed with a learning rate of 0.0001.
This suggests that larger networks with more hidden units may require smaller learning rates to train effectively. Moreover, while increasing the hidden state dimension can reduce errors and narrow the distribution spread, especially for MAPE, the best overall results in terms of both MAPE and angular score are achieved with 20 hidden units.

\subsubsection{Stacked LSTMs}
In our context, adding more LSTM layers does not improve model performance. For most configurations, a single LSTM layer provides better or comparable performance to deeper models. For example, in a setup with 20 hidden units, a sequence size of 30, and a learning rate of 0.0001, increasing the number of stacked LSTMs from 1 to 5 increases the MAPE from 15.72\% to 20.87\% and the angular score from 3.97° to 4.90°.

\subsubsection{Learning rate}
The experiments indicate that a learning rate of 0.001 generally leads to better results compared to a rate of 0.0001. For instance, in a configuration with a sequence size of 20, 10 hidden units, 5 LSTM layers, and a learning rate of 0.001, the model achieves a MAPE of 13.46\% and an angular score of 4.70°. In contrast, with a learning rate of 0.0001, the same configuration results in a MAPE of 34.75\% and an angular score of 8.24°.

Based on these observations, the hyperparameters of our model are set as follows: the sequence size is set to 10, the hidden state dimension to 20, the number of LSTM layers to 1, and the learning rate to 0.001. Although the lowest MAPE is achieved with 5 LSTM layers, this result is not consistent with other configurations using the same number of layers. Additionally, a sequence size of 10 yields the best results for the angular score, which is crucial for achieving the precision required in this regression task.

\begin{table}[htbp]
\small
\centering
\setlength{\tabcolsep}{1pt}
\begin{tabularx}{\linewidth}{|Y|Y|Y|Y|Y|Y|}
\hline
Sequence size & Hidden state dimension & Stacked LSTMs & Learning rate & MAPE (\%) & Angular score (°) \\
\hline
10 & 10 & 1 & 0.001 & 11.10 & 2.81 \\
10 & 10 & 1 & 0.0001 & 29.51 & 3.41 \\
20 & 10 & 1 & 0.001 & 17.70 & 4.57 \\
20 & 10 & 1 & 0.0001 & 30.20 & 5.27 \\
30 & 10 & 1 & 0.001 & 11.67 & 3.70 \\
30 & 10 & 1 & 0.0001 & 19.10 & 4.31 \\
10 & 20 & 1 & 0.001 & 10.52 & \textbf{2.77} \\
10 & 20 & 1 & 0.0001 & 17.07 & 3.04 \\
20 & 20 & 1 & 0.001 & 15.63 & 4.57 \\
20 & 20 & 1 & 0.0001 & 24.86 & 5.15 \\
30 & 20 & 1 & 0.001 & 12.68 & 3.72 \\
30 & 20 & 1 & 0.0001 & 15.72 & 3.97 \\
10 & 100 & 1 & 0.001 & 12.02 & 4.00 \\
10 & 100 & 1 & 0.0001 & 11.93 & 2.83 \\
20 & 100 & 1 & 0.001 & 14.01 & 4.61 \\
20 & 100 & 1 & 0.0001 & 12.77 & 4.60 \\
30 & 100 & 1 & 0.001 & 12.40 & 3.95 \\
30 & 100 & 1 & 0.0001 & 13.51 & 3.78 \\
10 & 10 & 5 & 0.001 & 24.25 & 3.62 \\
10 & 10 & 5 & 0.0001 & 34.06 & 5.66 \\
20 & 10 & 5 & 0.001 & 13.46 & 4.70 \\
20 & 10 & 5 & 0.0001 & 34.75 & 8.24 \\
30 & 10 & 5 & 0.001 & 18.50 & 4.04 \\
30 & 10 & 5 & 0.0001 & 37.45 & 7.21 \\
10 & 20 & 5 & 0.001 & 17.33 & 4.19 \\
10 & 20 & 5 & 0.0001 & 25.53 & 4.71 \\
20 & 20 & 5 & 0.001 & 15.53 & 4.60 \\
20 & 20 & 5 & 0.0001 & 26.64 & 6.35 \\
30 & 20 & 5 & 0.001 & \textbf{8.83} & 3.94 \\
30 & 20 & 5 & 0.0001 & 20.87 & 4.90 \\
10 & 100 & 5 & 0.001 & 21.10 & 4.00 \\
10 & 100 & 5 & 0.0001 & 15.96 & 2.88 \\
20 & 100 & 5 & 0.001 & 18.71 & 4.87 \\
20 & 100 & 5 & 0.0001 & 15.33 & 4.79 \\
30 & 100 & 5 & 0.001 & 12.07 & 4.04 \\
30 & 100 & 5 & 0.0001 & 17.48 & 4.10 \\
\hline
\end{tabularx}
\caption{Proposed model's performance for different learning hyperparameters}
\label{tab:tuning}
\end{table}

\subsection{Comparison with baseline predictors}
\label{evaluation_comparison}

Table \ref{tab:model_comparison} presents a comparative evaluation of the proposed LSTM model against several baseline machine learning models commonly used in the literature. The selected models are Multi-Layer Perceptron (MLP), ResNet 18 \cite{he_deepresiduallearning_2015}, Convolutional Neural Network (CNN) \cite{lecun_gradientbasedlearningapplied_1998}, Transformer \cite{vaswani_attentionallyou_2023} and mLSTM \cite{krause_multiplicativelstmsequence_2017}. 

To ensure a fair comparison, hyperparameters shared with our LSTM model are set to the same values. These included the learning rate for all models, sequence size and number of layers for Transformers and mLSTM, and hidden units for mLSTM.

Results show that models designed for sequence processing (LSTM, mLSTM, and Transformers) outperform the other models (MLP, ResNet 18, and CNN) in predicting wave direction. Specifically, mLSTM and Transformers models achieve MAPE values of 11.66\% and 11.70\%, respectively, and angular scores of 2.85° and 2.86°. While these two models perform well, the proposed LSTM architecture achieves slightly better results, with a MAPE of 10.52\% and an angular score of 2.77°.

\begin{table}[htbp]
\centering
\caption{Comparison of ML-based models for wave direction prediction}
\label{tab:model_comparison}
\begin{tabular}{|l|c|c|}
\hline
\textbf{Model} & \textbf{MAPE (\%)} & \textbf{Angular score (°)} \\
\hline
MLP & 34.35 & 5.27 \\
\hline
Transformer \cite{vaswani_attentionallyou_2023} & 11.66 & 2.85 \\
\hline
ResNet 18 \cite{he_deepresiduallearning_2015} & 28.43 & 5.67 \\
\hline
CNN \cite{lecun_gradientbasedlearningapplied_1998} & 33.42 & 5.23 \\
\hline
mLSTM \cite{krause_multiplicativelstmsequence_2017} & 11.7 & 2.86 \\
\hline
\textbf{Proposed model} & \textbf{10.52} & \textbf{2.77} \\
\hline
\end{tabular}
\end{table}

\subsection{Evaluation on unseen data}
\label{evaluation_sea_data}

The proposed model is evaluated using sea data that were not included in the training phase to assess its generalization capabilities. A total of 136,041 raw data points are collected across six trajectories. Each trajectory lasts between 6 and 14 minutes. After applying the pre-processing steps described in Section \ref{model_overview_data_processing}, the dataset is reduced to 134,350 data points.

The predicted dominant wave direction has a mean value of 314.64°. As noted in Section \ref{evaluation_ground_truth}, the wave direction was assessed using fixed landmarkers, resulting in an estimated direction of 335° with a tolerance of ± 5°. Based on this estimation, the model's average error is determined to be 20.36°.

\begin{figure}[ht]
    \centering
    \includegraphics[width=0.7\linewidth]{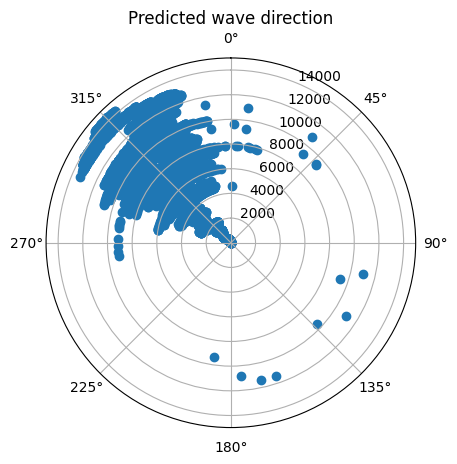}
    \caption{Predicted wave direction for open sea data - First trajectory}
    \label{fig:prediction_sea_first}
\end{figure}

\begin{figure*}[ht]
    \centering
    \includegraphics[width=\linewidth]{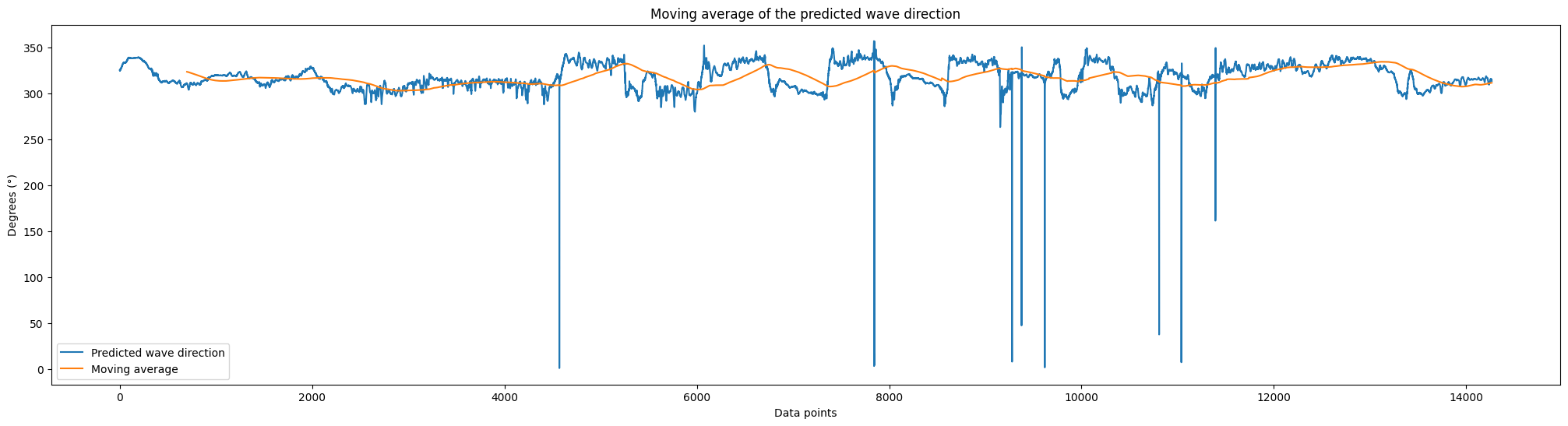}
    \caption{Moving average on predicted wave direction for open sea data - First trajectory}
    \label{fig:moving_average_prediction_sea_first}
\end{figure*}

Figure \ref{fig:prediction_sea_first} illustrates the predicted wave direction for the first trajectory (after subtracting the drone's yaw). Applying a moving average with a window of approximately 16 seconds reduces the standard deviation to 13.78°, resulting in a notably more stable prediction, as depicted in Figure \ref{fig:moving_average_prediction_sea_first}.

\section{Related work}
\label{related_work}
In \cite{mittendorf_seastateidentification_2022}, the authors aim to predict sea state parameters (significant wave height, peak period, and mean encounter wave direction) using data collected from a container vessel navigating the Northern Atlantic over a 1.5-year period. Ground truth is established through a directional wave radar. Pre-processing involves combining wave radar data with other sensor readings, including an MRU (Motion Response Unit) capturing six degrees of freedom, and GPS data. The dataset is filtered to focus on deep sea conditions, which are deemed more appropriate given the ship's large size. Several models are evaluated for this regression task, including ResNet (Residual Network), Inception Network, MLSTM (Multi-Channel Convolutional LSTM), and CNN (Convolutional Neural Network). Training is conducted on data sequences from both the temporal and frequency domains. The results show that models using the frequency domain generally outperforms those in the time domain in both prediction accuracy and computational efficiency, with the Inception Network achieving the highest performance. However, the frequency domain approach is limited by its reliance on relatively long time windows (25 to 30 minutes), which, while manageable for ships, might restrict the flexibility and responsiveness of models in operational settings for more dynamic platforms such as drones.

In \cite{cheng_modelinganalysismotion_2019}, the authors introduce the SeaStateNet model, which integrates LSTM, CNN, and FFT (Fast Fourier Transform) with feature fusion to classify sea states based solely on wave height.

In \cite{cheng_noveldenselyconnected_2020}, ship motion data are used to classify sea states based on wave height and direction. The authors categorize sea states into 40 unique classes using a combination of sea state codes and directional intervals. Their model relies on stacked CNNs with dense connections and feature attention mechanisms.

In \cite{han_datadrivenseastate_2021}, a more traditional approach is taken with the use of models such as SVR (Support Vector Regression), KNN (K-Nearest Neighbors), and GBDT (Gradient-Boosted Decision Trees). The authors generate features through statistical, temporal, spectral, and wavelet analyses to enhance the predictive accuracy of the models.

\section{Conclusion}
\label{conclusion}
In this paper, a predictive model leveraging an LSTM architecture is proposed to estimate wave direction using data from a drone's navigation and orientation sensors. The model is evaluated through experiments conducted in a controlled experimental pool and a real-world open sea environment. The results highlight the model's potential to improve the navigation and operational safety of USVs in dynamic marine settings.

To further refine wave direction estimation, reliable field measurements of wave direction are needed (e.g., through the processing of images captured by an aerial drone). These measurements should be performed for each wave encountered by the drone, as the direction can fluctuate and evolve over time, in order to precisely measure the model's error.

Future work will also focus on expanding the dataset through further experiments in varied sea conditions and incorporating other maritime variables such as wind and current. This will help capture a broader range of scenarios and refine the model's accuracy. Additionally, we plan to adapt the model for real-time applications, addressing challenges such as optimizing energy consumption and minimizing inference times. We also intend to combine our wave direction prediction model with a reinforcement learning agent to simulate and assess the USV's behavior in complex and hostile environments, enhancing the robustness and autonomy of its navigation capabilities.

\bibliographystyle{IEEEtran}
\bibliography{references}

\end{document}